\documentclass{article}

     \PassOptionsToPackage{numbers, compress}{natbib}

\usepackage[final]{neurips_2020_ml4ad}

\usepackage[utf8]{inputenc} 
\usepackage[T1]{fontenc}    
\usepackage{hyperref}       
\usepackage{url}            
\usepackage{booktabs}       
\usepackage{amsfonts}       
\usepackage{nicefrac}       
\usepackage{microtype}      
\usepackage{times}
\usepackage{epsfig}
\usepackage{graphicx}
\usepackage{amsmath}
\usepackage{amssymb}
\usepackage{bbm}
\usepackage{xcolor}
\usepackage{multirow}

\title{MODETR: Moving Object Detection with Transformers}

\author{%
Eslam Mohamed \\
Deep Learning Research  \\
Valeo R\&D Cairo, EGYPT \\
\texttt{eslam.mohamed-abdelrahman@valeo.com} \\
  \And
  Ahmad El Sallab \\
  Deep Learning Research \\
  Valeo R\&D Cairo, EGYPT \\
  \texttt{ahmad.el-sallab@valeo.com} \\
}

\begin{document}

\maketitle

\begin{abstract}

 Moving Object Detection (MOD) is a crucial task for the Autonomous Driving pipeline. MOD is usually handled via 2-stream convolutional architectures that incorporates both appearance and motion cues, without considering the inter-relations between the spatial or motion features.  In this paper, we tackle this problem through multi-head attention mechanisms, both across the spatial and motion streams. We propose MODETR; a Moving Object DEtection TRansformer network, comprised of multi-stream transformer encoders for both spatial and motion modalities, and an object transformer decoder that produces the moving objects bounding boxes using set predictions. The whole architecture is trained end-to-end using bi-partite loss. Several methods of incorporating motion cues with the Transformer model are explored, including two-stream RGB and Optical Flow (OF) methods, and multi-stream architectures that take advantage of sequence information. To incorporate the temporal information, we propose a new Temporal Positional Encoding (TPE) approach to extend the Spatial Positional Encoding (SPE) in DETR. We explore two architectural choices for that, balancing between speed and time. To evaluate the our network, we perform the MOD task on the KITTI  MOD \cite{siam2018modnet} data set. Results show significant 5\% mAP of the Transformer network for MOD over the state-of-the art methods. Moreover, the proposed TPE encoding provides 10\% mAP improvement over the SPE baseline.
\end{abstract}

\section{Introduction}
Identifying static and dynamic objects in the scene is crucial for the mapping and planning tasks in the Autonomous Driving pipeline, especially in highly dynamic scenes. This motivates the Moving Object Detection (MOD) perception task.

Motion cues are of particular importance to MOD, more than in traditional object detection tasks, where we want to localize and classify the object category. In MOD, the object class is its motion type: Moving vs. Static. The motion can be due to other dynamic objects, or due to the ego vehicle itself, which introduces a relative motion than might incorrectly lead to perceiving static objects as moving. This adds more complexity to the motion classification task, and poses an interesting question of how to represent the objects motion.

To address this question, several approaches are explored. Optical Flow (OF) motion cues were explored in \cite{siam2018modnet}, in addition to spatial RGB frames. OF has downsides of complex computation in AD systems, in addition to including the ego-motion itself which hurts the perception of static objects as moving \cite{ramzy2019rst}. However, modern hardware platforms, like Nvidia Xavier, TI TDA4x and Renesas V3H, have dedicated chips for calculating OF, which reduces the overhead. Recurrent sequence models (ConvLSTM) are explored in \cite{ramzy2019rst}. However, when it comes to AD systems deployment, sequential recurrent models like LSTM may incur extra time due to their sequential nature and their Auto-Regression (AR) decoding process \cite{vaswani2017attention}.

Recently, Transformer networks are applied in the computer vision domain for object detection in DETR \cite{carion2020end}, following their success in language domain \cite{vaswani2017attention}. While Transformers are originally proposed in the context of sequence-to-sequence tasks like Neural Machine Translation, which lend itself naturally to be extended to temporal modeling, only their spatial attention side is exploited in \cite{carion2020end} for object detection. 

In this work, we explore the incorporation of the Transformer in the task of MOD. We adopt a 2-stream architecture as in  \cite{siam2018modnet}, extending the Transformer Encoder-Decoder architecture; DETR \cite{carion2020end}. We first try RGB+OF 2-stream architecture, and let the DETR handles the inter-relations across the features, both in spatial (RGB) and motion (OF) streams. Then we try RGB+RGB architectures that tries to include the motion information from 2 consecutive frames. We explore different ways of including the temporal aspect in the DETR architecture, by introducing Temporal Positional Encoding (TPE), in addition to the Spatial Positional Encoding (SPE) in two architectures, that balance between speed and performance.

The contributions of this papers can be listed as follows:
\begin{itemize}
    \item MODETR: Modifying the DETR architecture into a 2-stream architecture to perform MOD.
    \item Bench marking the performance of DETR with different motion cues OF and stacked RGB frames.
    \item Modifying DETR to be time-aware, through Temporal Positional Encoding (TPE).
    
\end{itemize}

For evaluation, we use the published dataset KITTI MOD \cite{siam2018modnet} which includes the motion masks. We benchmark our models against convolutional architectures, as in MODNet \cite{siam2018modnet} and YOLACT \cite{bolya2019yolact}. Results show 10\% in mAP over the baseline RGB-only architecture, adn around 5\% over the best state-of-the art MOD approaches. Also, the proposed Temporal Positional Encoding (TPE) provides 2\% mAP improvement, which paves the way for future spatio-temporal models that accounts for multi-step sequences.

\begin{figure*}
\begin{center}
\centerline{\includegraphics[width=150mm]{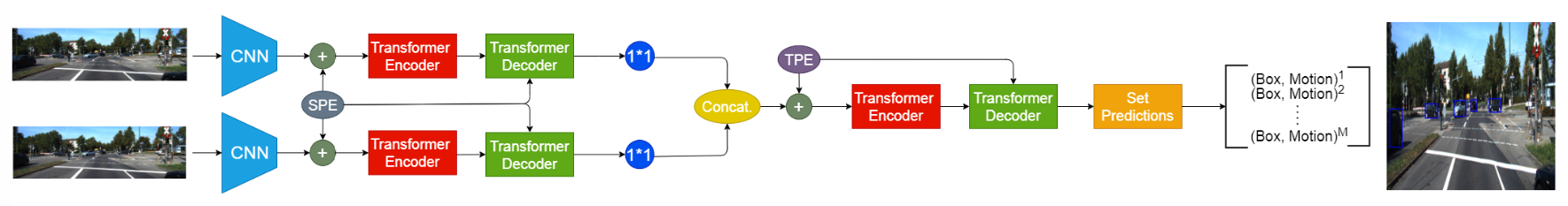}}
\caption{\label{MODETR} Moving Object DEtection TRansformer (MODETR) Architecture}
\label{icml-historical}
\end{center}
\end{figure*}

\section{Approach}

The general multi-stream MODETR architecture is shown in \ref{MODETR}. In general MODETR has the following components: \textbf{CNN backbone}: which encodes the spatial or motion stream features. \textbf{DETR Encoder}: which comprises both Transformer encoder and decoder layers. Each stream undergoes multi-head self attention operation, using Spatial Poisition Encoding $SPE$ as in \cite{carion2020end}. \textbf{Fusion}: which merges the streams features, using concatenation or 1x1 convolution. \textbf{DETR Decoder}: which is responsible of generating the output features, based on multi-head attention over the fused streams features, and the learned object queries as in \cite{carion2020end}. The \textbf{Set predictions}: to produce the final object predictions, as the bounding box parameters, and the moving/static classification. The MODETR shall predict up to $N$ set predictions of moving objects, following the bi-partite matching algorithm in \cite{carion2020end}, which enables end-end training, without any post-processing.

MOD task depends on feeding both motion and appearance or spatial features. Appearance cues are present in the spatial RGB frames features. We explore two setting to include motion: 1) Two-stream RGB, referred to as RGB+RGB in our experiments, and 2) Optical flow motion cuses, which we refer to as RGB+OF. We study and discuss both setups in details, with the architectural modifications over the generic MODETR in Figure\ref{MODETR}.

\subsection{Two-stream RGB (RGB+RGB)}

In this setup we explore the possibility of learning the motion cues from consecutive RGB frames. Each frame shall pass by a CNN backbone, which extracts a features map of size $H \times W \times C$, where $H \times W$ is the frame size, and $C$ is the number of features. This 3D tensor is then flattened to $H*W \times C$ 2D tensor, which enables the Spatial Positional Encoding (SPE) to attend to each pixel differently, and capture the spatial relations. The SPE range shall be $SPE \subset [1, H*W]$.

We propose to add Temporal Position Encoding (TPE) to incorporate frame sequence information in the Encoder-Decoder DETR architecture \cite{carion2020end}. Similar to Positional Encoding in \cite{vaswani2017attention}, we want our TPE to index the time frames in the same way the words sequence are indexed in the Positional Embedding block of the Transformer Encoder. This will enable the Encoder to to attend to different spatial frames at different time steps within a certain window. Since we do not have a need to decode multiple frames, there is no need to add TPE at the Decoder side. The TPE Embedding block, will have input indices $TPE \subset [1,N]$, according to the frame order in the window of $N$ frames, which 2 in our case.

\subsection{Optical flow motion stream (RGB+OF)}

This architecture is similar to RGB+RGB setup, except that we feed to each stream DETR Encoder the motion information through the Optical Flow (OF) map highlighting pixels motion. In this approach, we make use of FlowNet 2.0~\cite{Ilg2016FlowNet2E} model to compute optical flow. The fusion between appearance (RGB) and motion (OF) is performed on the feature level. This architecture has the advantage of performing self-attention both across spatial and motion features, which encodes the cross relations in both modalities. 

\begin{figure*}
\begin{center}
\centerline{\includegraphics[width=150mm]{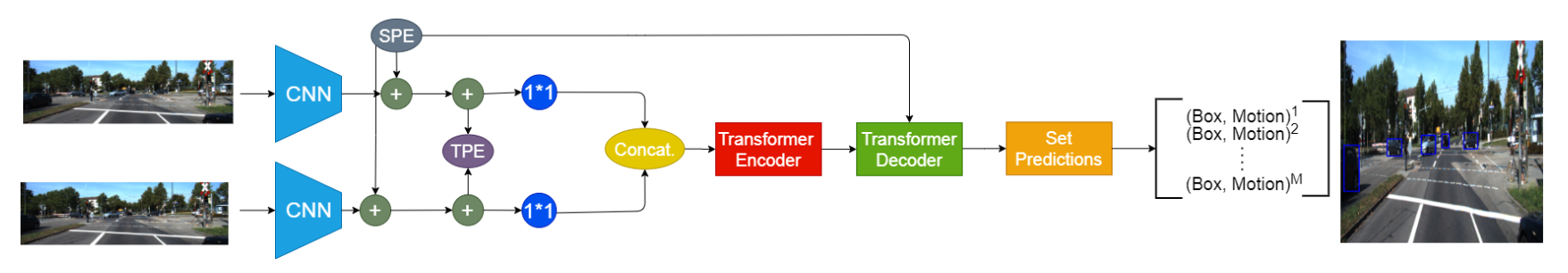}}
\caption{\label{Early} 2-stream RGB+RGB with early TPE}
\label{icml-historical}
\end{center}
\end{figure*}
\section{Experimental setup}

\textbf{RGB-only Baseline.} A single RGB image was fed to ResNet50 backbone to extract spatial feature from the image followed by an encoder-decoder transformer, followed by prediction head that produces detected objects. In this setup a spatial positional encoding was used to encode the pixel location information.

\textbf{RGB+RGB} Two successive frames go through a shared backbone and transformer's encoder that are described in the baseline case, followed by $1 \times 1$ conv block to reduce the feature map size by half on channel dimension to reduce model complexity, then concatenate both features that are extracted from the two consecutive frames and feed them to transformer's decoder. No TPE in this setup.

The Multi-stream nature of the input introduces different architectural choices. We explore two different architectures:

\textbf{RGB+RGB (Early TPE)}: In this setup, the TPE Embedding is simply added to the encoded CNN features, together with the SPE Embedding. In this case, the multiple streams paths are reduced to just the CNN encoded spatial features, with the SPE and TPE embeddings added, followed by the DETR Decoder, as shown in Figure \ref{Early}. This has the advantage of using a common Transformer Encoder-Decoder architecture, which saves the inference time and memory. However, having simple addition of different space and time encoding might weaken the representation power of the frame features vector, and make the Transformer Encoder task harder to model both aspects of space and time.

\begin{figure*}
\begin{center}
\centerline{\includegraphics[width=150mm]{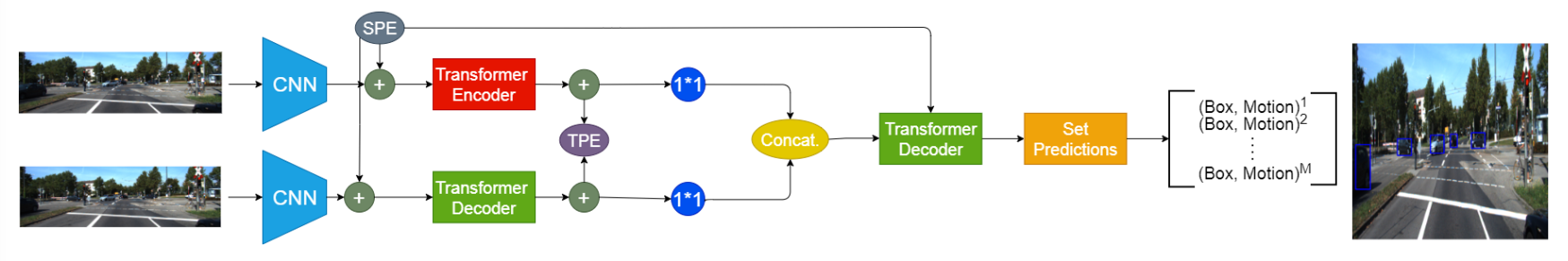}}
\caption{\label{Late} 2-stream RGB+RGB with late TPE}
\label{icml-historical}
\end{center}
\end{figure*}

\textbf{RGB+RGB (Late TPE)}: In this setup, each frame will have its own Transformer Encoder, which takes care of the spatial features, as shown in Figure \ref{Late}. Following, TPE block will differentiate each frame representation according to its order in the frames sequence. This architecture suffers from slow training and inference time, and larger model size, where we have $N$ Transformer Encoders. On the other hand, it leverages the representation power of the spatial aspect, before adding the temporal embedding, in a hierarchical fashion.

\subsection{Results and Discussion}

\begin{table*}[t]
\label{tab:MODETR}
\begin{center}
\begin{tabular}{l|c|c|c}
\hline
Method & $mAP_{Total}$ & $mAP_{50}$ & $mAP_{75}$ \\
\hline
RGB-only Baseline              & 23\% & 42.2\% & 23.7\%                          \\
RGB+RGB           & 25.3\% & 47.2\% & 24.5\%                        \\
RGB+RGB (Early TPE) & 25.3\% & 47.86\% & 24.84\%    \\
RGB+RGB (Late TPE) & 24.4\% & 45.6\%  & 24.1\%     \\
RGB + Optical Flow    & \textbf{33.9\%} & \textbf{59.3\%} & \textbf{37.2\%}         \\
\hline
\end{tabular}
\end{center}
\caption{MODETR results of two-stream DETR architectures}
\end{table*}

\begin{table*}[t]
\label{tab:MODETR_CONV}
\begin{center}
\begin{tabular}{l|c|c|c}
\hline
Method & $mAP_{Total}$ & $mAP_{50}$ & $mAP_{75}$ \\
\hline
\hline
RGB-only Baseline              & 29.2\% & 48.3\% & 32.1\% \\ 
\hline
\multicolumn{4}{c}{\textbf{MODETR}} \\
\hline
                        \\
RGB+RGB (Early TPE) & 33.3\% & 53.2\%  & 38.2\%     \\
RGB + Optical Flow    & \textbf{42.9\%} & \textbf{66.3\%} & \textbf{50.9\%}         \\
\hline
\multicolumn{4}{c}{\textbf{Convolutional 2-stream}} \\
\hline
                        \\
RGB+RGB & 17.9\% & 36.86\%  & 14.73\%     \\
MODNet \cite{siam2018modnet} (RGB + OF)    & 32.04\% & 61.6\% & 29.47\%         \\
InstanceMotSeg \cite{mohamed2020instancemotseg} (RGB+OF)    & 40.6\% & 59\% & 49\%         \\
\hline
\end{tabular}
\end{center}
\caption{Comparison of MODETR vs Convolutional 2-stream models }
\end{table*}

As shown in Table \ref{MODETR}, results are highly in favor of RGB+OF model, which is somehow expected due to the explicit motion features extracted with Flownet. This comes at the expense of increased processing time as discussed before. In terms of implicit motion cues: RGB+RGB architectures, the results show superior performance of the early TPE architecture. From one hand, the temporal information through TPE is preserved within the Encoder Transformer, which enables the attention mechanism to capture the temporal relations. On the other hand, the limited window horizon $N=2$ in our experiments, enables the Encoder to capture both the temporal and spatial positions across the 2 frames. We expect the performance to deteriorate with the increased window size, which is left to future work. In this experiment we use image resolution $480 \times 145$.

\begin{figure*}
\begin{center}
\centerline{\includegraphics[width=\textwidth]{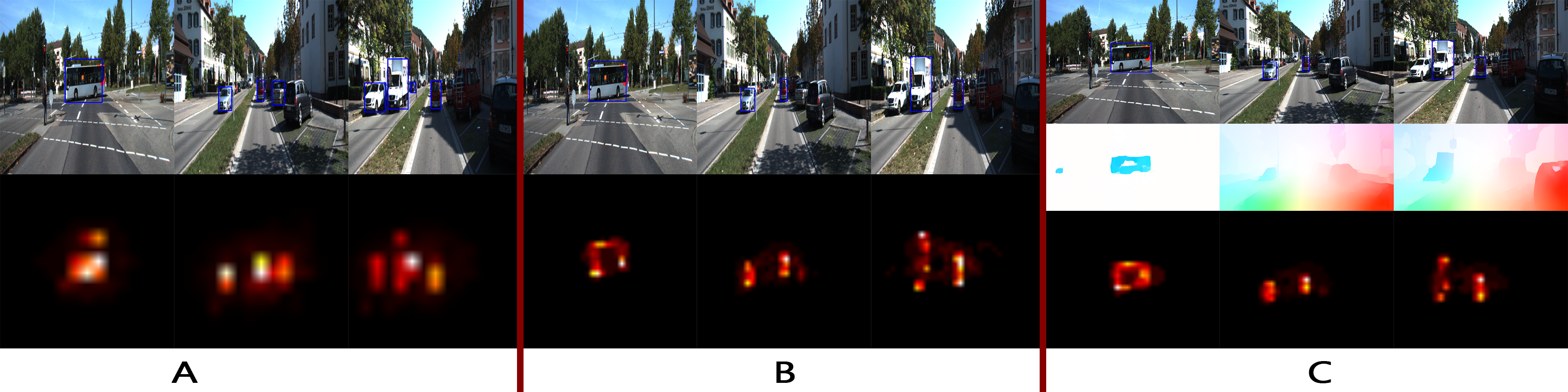}}
\caption{Results for MODETR-Baseline, MODETR-2RGBs and MODETR-RGB+OF: The first two rows show the visual outputs and the attention maps for baseline, the third and fourth rows show the visual outputs and the attention maps for 2RGBs and the last three rows show the visual outputs, corresponding optical flow and the attention maps for RGB+OF}
\label{results}
\end{center}
\end{figure*}

We also inspect the effect of OF and Attention maps. Visual samples are shown in Figure \ref{results}. In general, the attention maps goes to the moving objects as expected, and help focusing the final box away from static objects. This is clear even in the RGB-only baseline. The effect of 2-stream architectures starts to appear from the RGB+RGB architecture, where attention begins to focus on the object boundaries, where the motion is most clear. With the introducion of RGB+OF, the network pays more attention to the moving parts of the object, where the wheels of the bus receives more attention. This is also linked to the OF stream, which is visualized in the RGB+OF row.

We benchmark against state-of-the art motion detection approaches: 1) MODNet \cite{siam2018modnet} and 2) InstanceMotSeg  \cite{bolya2019yolact}. Again, the 2-stream MODETR with RGB+OF outperforms both baselines. It is worth mentioning that, for networks that perform multi tasks, as in MODNet \cite{siam2018modnet}, we focus our comparison to the detection head, which is aligned to our MOD task, disregarding the motion segmentation masks. This adds a point in favor of DETR, since the common encoder in such approaches is expected to benefit from the data from both tasks, however, MODETR is able to beat them. As future work, Multi-Task Learning (MTL) can be explored with MODETR, which is expected to further improve. In this experiment we use image resolution $550 \times 550$, in order to be able to compare to InstanceMotSeg \cite{mohamed2020instancemotseg} and MODNet \cite{siam2018modnet}.

\section{Conclusion}
In this paper we presented MODETR, a 2-stream Transformer based network for MOD. We explored different architectures to extend the basic DETR network to handle the temporal aspect for the task of MOD. We presented a novel method to handle multi-step sequential frames, through Temporal Positional Embedding, which is a step towards Spatio-Temporal extension of the basic DETR model. Early embedding shows better performance, due to the limited window horizon in our experiments. The employment of explicit motion features through OF, together with appearance features through the spatial RGB raw frames, produce the best results.

{\small
\bibliographystyle{ieee_fullname}
\bibliography{egbib}
}

\end{document}